\documentclass[runningheads,table]{llncs} 

\usepackage{eccv}

\usepackage{eccvabbrv}

\usepackage{graphicx}
\usepackage{booktabs}

\usepackage[accsupp]{axessibility}  

\usepackage{colortbl}
\usepackage{multirow}
\usepackage{array}
\usepackage{svg}
\usepackage{amsmath}
\usepackage{amssymb}
\usepackage{pifont}
\usepackage{dsfont}
\usepackage{wrapfig}
\usepackage{mathtools}
\usepackage{enumitem}
\usepackage[dvipsnames]{xcolor}

\usepackage{hyperref}
\hypersetup{
    colorlinks=true,
    linkcolor=BrickRed,
    filecolor=Fuchsia,      
    urlcolor=Orchid,
    citecolor=eccvblue,
}

\usepackage{orcidlink}

\newcommand{\myparagraph}[1]{\vspace{0.05cm}\noindent\textbf{#1}}
\definecolor{myrowcolor}{rgb}{0.9, 0.9, 0.9}
\newcommand{\ret}[2]{#1$\rightarrow$#2}
\definecolor{Fbestcolor}{HTML}{b8f7ac}
\definecolor{Sbestcolor}{HTML}{e2ffda}
\newcommand{\Fbest}[1]{\cellcolor{Fbestcolor} #1}
\newcommand{\Sbest}[1]{\cellcolor{Sbestcolor} #1}

\begin{document}

\title{PoseEmbroider: Towards a 3D, Visual, Semantic-aware Human Pose Representation}

\titlerunning{PoseEmbroider}

\author{Ginger Delmas\inst{1,2} \and
Philippe Weinzaepfel\inst{2} \\
Francesc Moreno-Noguer\inst{1} \and
Gr\'egory Rogez \inst{2}}


\authorrunning{G.~Delmas et al.}

\institute{Institut de Robòtica i Informàtica Industrial, CSIC-UPC, Barcelona, Spain\\
\and
NAVER LABS Europe\\[0.2cm]
{{\small \url{https://europe.naverlabs.com/research/PoseEmbroider/}}\normalsize}
}

\maketitle

\begin{abstract}
Aligning multiple modalities in a latent space, such as images and texts, has shown to produce powerful semantic visual representations, fueling tasks like image captioning, text-to-image generation, or image grounding.
In the context of human-centric vision, albeit CLIP-like representations encode most standard human poses relatively well (such as standing or sitting), they lack sufficient acuteness to discern detailed or uncommon ones. 
Actually, while 3D human poses have been often associated with images (\eg to perform pose estimation or pose-conditioned image generation), or more recently with text (\eg for text-to-pose generation), they have seldom been paired with both.
In this work, we combine 3D poses, person's pictures and textual pose descriptions to produce an enhanced 3D-, visual- and semantic-aware human pose representation.
We introduce a new transformer-based model, trained in a retrieval fashion, which can take as input any combination of the aforementioned modalities. 
When composing modalities, it outperforms a standard multi-modal alignment retrieval model, making it possible to sort out partial information (\eg image with the lower body occluded). 
We showcase the potential of such an embroidered pose representation for (1) SMPL regression from image with optional text cue; and (2) on the task of fine-grained instruction generation, which consists in generating a text that describes how to move from one 3D pose to another (as a fitness coach).
Unlike prior works, our model can take any kind of input (image and/or pose) without retraining.

\end{abstract}

\section{Introduction}
\label{sec:intro}

People play a central role in many applications across a wide range of domains, including robotics, digitization (such as virtual avatars), and entertainment. In many of these contexts, the human pose is a defining characteristic.
While a large body of work aims to estimate~\cite{smplify, goel2023humans, kolotouros2019learning} or predict it~\cite{posegptnaver, barsoum2018hp, Ahuja2019Language2PoseNL}, for instance, to further facilitate human-robot interaction, another seeks to generate it~\cite{plappert2016kit, petrovich2022temos, Ghosh_2021_ICCV, Guo_2022_CVPR, kim2022flame}, to enhance experiences in video games or virtual worlds. These tasks demonstrate the crucial importance of human understanding.

Early works have focused on detecting and visually understanding people. While human bodies can already be fairly well studied through visual data, true human understanding goes beyond mere perception.
It also relies on \textit{meaning}, that is, semantics. Now, we humans, tend to prefer when the world's semantics match ours. This is where natural language comes into play.
Language empowers the conveyance of complex and abstract concepts; making it possible to gather similar elements together under the same word. For instance, one person could have their hand at shoulder level, and another person their hand way overhead; yet, both individuals could be ``\emph{waving}''.

\begin{figure}[t]
    \centering
    \includegraphics[width=\linewidth]{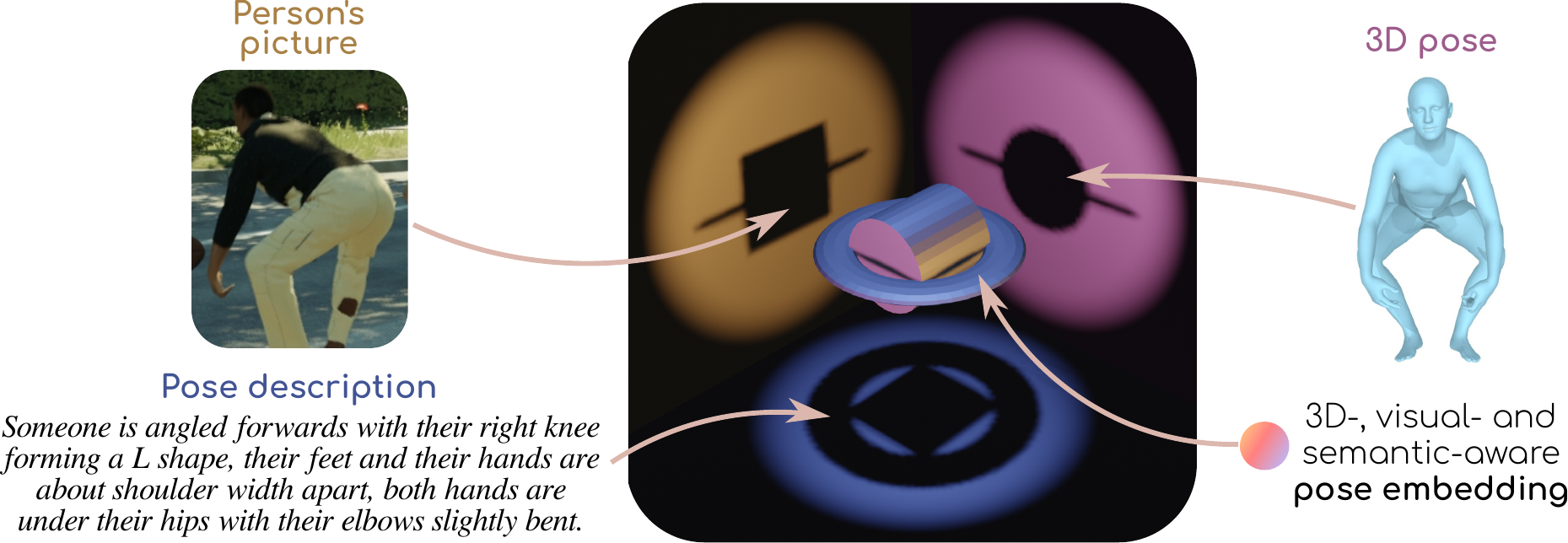} \\[-0.25cm]
    \caption{\textbf{Motivation.} Comprehending a complex 3D object in a 2D world is not simple. Having access to several of its shadows, obtained by lighting it under different angles, can help better understand it. Similarly, we collect several multi-modal (and naturally partial) observations of the human pose (the ``shadows''), and try to create an enriched pose embedding (the ``3D'' object). This embedding is derived from 3D joint rotations, pictures of humans and pose descriptions, then further used in downstream applications requiring human pose understanding.} 
    \label{fig:motivation_illustration}
\end{figure}

Ultimately, both visual and textual data are essential to achieve human understanding: they are two facets of the same prism. However, both are imperfect: visual data may exhibit occlusions or depth uncertainty, while text is relatively ambiguous. Despite these flaws, they provide crucial information that a 3D pose alone could not convey, such as world affordance, reality anchoring, and semantics.
In the end, all three modalities (visual data, text and 3D poses) can be considered complementary -- partial, yet valuable -- observations of the same abstract ``human pose'' concept (see Figure~\ref{fig:motivation_illustration} for an illustration).

More concretely, recent advances have demonstrated the utility of pairing images and texts to derive powerful semantic image embeddings~\cite{clip}. In this work, we extend this principle to the concept of human pose. We aim to derive a rich pose embedding that is simultaneously semantic-, visual- and 3D-aware, by embroidering images, texts and 3D poses together. Indeed, current endeavors only yield coarse representations of human poses, failing to distinguish between two similar complex poses. 

Previous works have essentially focused on connecting individuals depicted in images to their 2D or 3D pose~\cite{xu2022vitpose, goel2023humans, chen2022liftedcl, yuan2024hap}, or on linking 3D poses with fine-grained text descriptions~\cite{posescript}, thereby producing strong visual pose embeddings \emph{or} semantic pose embeddings. More recently \cite{feng2024chatpose} repurposed a large vision and language transformer model to output a human pose, based on either an image, text or a combination of both inputs. However, just like other multi-modal works leveraging large language models (LLM)~\cite{wu2023next, driess2023palme}, it requires converting new modalities to equivalent textual representations, so as to enable processing in the LLM space. This process could lead to partial loss of modality-specific information, in particular information that cannot be transcribed through text. Another body of works~\cite{wang2023hulk, ci2023unihcp, tang2023humanbench} proposes to unify human-centric perception tasks under a single multi-modal model. Yet, these models are generally trained with task-specific objectives. 
Overall, recent multi-modal methods tend to \textit{align} modalities to enable any-to-any translation~\cite{girdhar2023imagebind, mizrahi20244m}, but do not necessarily combine multi-modal information to build a single versatile representation.

In this paper, we design a multi-modal framework that \textit{embroiders} different modalities, so as to build a richer semantic-, visual-, 3D-aware pose embedding space. We use a transformer to aggregate information from available modalities within a single global token. The model is trained with uni-modal contrastive objectives, on the reprojections of this global representation to each modality space.
As a result, we can enhance any single modality embedding fed to our model with multi-modal awareness. 
We demonstrate the benefit of our proposed pose representation by addressing the tasks of any-to-any multi-modal retrieval, pose estimation, as well as pose instruction generation, which has a direct application in automatic fitness coaching. This task consists in producing a text that specifies how to modify one pose into another. 
Differing from the initially proposed baseline PoseFix~\cite{delmas2023posefix}, the utilization of our multi-modal representation makes it possible to process direct camera input without the need for additional retraining.
In summary, our contributions list as follow:
\begin{itemize}[topsep=0pt]
    \item[\ding{93}] We introduce a new framework to embroider together several human pose-related modalities and derive a rich semantic-, visual-, 3D-aware pose embedding space (Section~\ref{sec:main_method}), We train it on the adapted BEDLAM-Script dataset (a description-augmented version of BEDLAM~\cite{Black_CVPR_2023_bedlam}). As a direct side-product, we present results for any-to-any multi-modal retrieval (Section~\ref{sec:multimodal_retrieval}).
    \item[\ding{93}] We showcase an application of the proposed enhanced pose representation for the task of pose instruction generation (Section~\ref{sec:pose_instruct}). Although our method is almost exclusively trained on synthetic data (the proposed BEDLAM-Fix dataset), we obtain promising results on real-world images. 
    \item[\ding{93}] We illustrate SMPL regression as another application (Section~\ref{sec:pose_estimation}).
\end{itemize}

\section{Related Work}
\label{sec:related_work}

We propose a novel \emph{multi-modal human pose representation}, using a framework related to general \emph{multi-modal alignment}, which can be applied to downstream tasks such as \textit{pose instruction generation}.
We briefly review related methods.

\myparagraph{Multi-modal representations of humans.}
Several methods proposed to learn efficient pose-structured human image representations ~\cite{chen2022liftedcl,chen2023solider, yuan2024hap}, but do not consider valuable extraneous information (3D, or finer-semantics brought by text data).
A growing body of work, focusing on human-centric perception, use more human-related multi-modal data (\eg RGB images, depth maps, 2D keypoints, 3D pose and shape, visual attributes or description \etc)~\cite{hong2022hcmoco} to perform diverse tasks, \eg person re-identification, human parsing, pose estimation, action recognition, attribute recognition \etc.
These works usually resort to task- or dataset-specific objective functions, to train a single model taking multi-modal input. PATH~\cite{tang2023humanbench} learns a shared image transformer with task-specialized projectors and dataset-dedicated heads. UniHCP~\cite{ci2023unihcp} processes the image via cross-attentions taking task-specific queries, task-wise interpreted as a set of features directly denoting the expected task output. Less visual-centric, Hulk~\cite{wang2023hulk} trains modality-specific tokenizers which outputs are processed in a shared encoder and decoded based on given modality indicators (``query tokens'') to perform modality translation. 
Unlike these works, we introduce a task-agnostic multi-modal-aware representation, that could be used out-of-the box in any pose-related task. In particular, we show its effectiveness on the task of pose instruction generation from images, which requires adequate human pose perception and a fine-grained semantic understanding of the body/  parts and their relationships.

Closer to this idea, PoseScript~\cite{posescript} models semantic pose embeddings by pairing 3D poses and descriptions in natural language. However it does not consider the visual modality.
More recently, ChatPose~\cite{feng2024chatpose} append a SMPL~\cite{smpl} projection layer to a large vision and language model~\cite{liu2024visual}, so to leverage its reasoning abilities for pose estimation and text-to-pose generation.
They hence derive a visual-semantic 3D pose representation, yet constrained to live in the textual space. Also, the model is not designed to take direct 3D pose input.

\myparagraph{Multi-modal alignment.}
It is common to align different modalities to perform multi-modal applications.
Efforts spanned aligning text and images ~\cite{frome2013devise, kiros2014unifying, faghri2017vse, kwon2022maskedvlm}, videos~\cite{alayrac2020self, akbari2021vatt, Lin_2023_ICCV}, audio~\cite{Ibrahimi_2023_ICCV, muller2018cross}, robotic states~\cite{driess2023palme}, 3D shapes~\cite{ruan2024tricolo}, 3D scenes~\cite{Jin_2023_CVPR}, 3D human poses~\cite{posescript}, human motions~\cite{Petrovich_2023_ICCV, yin2024trimodal}, and so forth.
Beyond empowering cross-modal retrieval, connecting modalities gives birth to powerful multi-informed versatile encodings. One of the most recent iconic works is CLIP~\cite{clip}, which learns a joint embedding space for images and texts with contrastive learning. The produced visual semantic representations are reused off-the-shelf in a variety of tasks and domains~\cite{tevet2022motionclip, youwang2022clip, Kim_2022_CVPR, ding2024clip}.
Research in multi-modal alignment further stepped up thanks to the introduction of ever-growing datasets, computational resources and models, which make it possible to get qualitative pseudo labeling~\cite{xu2022vitpose, mizrahi20244m} or reliable synthetic data~\cite{brooks2023instructpix2pix}.

Some works have explored aligning more than two modalities. Omnivore~\cite{girdhar2022omnivore} aligns several unpaired labeled visual modalities by feeding all visual patches to a single transformer, trained for classification. All modalities end up being encoded in the same space, hatching cross-modal retrieval. ImageBind~\cite{girdhar2023imagebind} brings it one step further by additionally considering non-visual modalities such as audio and text, leveraging the natural co-occurrence of images with other modalities.

Other recent works align modalities to facilitate any-to-any generation. NExT-GPT~\cite{wu2023next}, similar to Palm-e~\cite{driess2023palme}, feeds pretrained uni-modal representations to learnable modality-specific projections layers, such that they can be processed by a frozen LLM~\cite{vicuna2023}.
Eventually, the outputs are re-projected in uni-modal spaces and fed to pretrained diffusion models for generation. 4M~\cite{mizrahi20244m} tokenizes all modalities to process them with the same tansformer decoder, trained with masked modeling (for visual modalities) and next token prediction (for sequence modalities), for a random subset of (modality) query and target tokens. It converts modalities but does not learn a single multi-modal-informed representation.

Different from the above-mentioned methods, we go beyond \textit{aligning} or \textit{converting} modalities. We learn to \textit{intermingle} them with the aim of obtaining a unique, richer, multi-modal-informed representation, computable from any set of input modalities. Basically, we use multi-modal data to figure how to \textit{enhance} uni-modal encodings. In particular, this augmented embedding is not compelled to live in the textual space as in~\cite{wu2023next, feng2024chatpose, driess2023palme}, which could lead to the loss of non-textual information. Instead, it is free to assume any relevant structure in its own embedding space. For the same reason, we apply contrastive learning on the modality-specific spaces instead of the augmented embedding space, using uni-modal reprojections of the augmented encoding.

\myparagraph{Pose instruction generation} is a recent task, which consists in generating an instruction explaining how to correct one pose in another specific pose.
FixMyPose~\cite{fixmypose} introduced a first dataset based on highly-synthetic pairs of images. AIFit~\cite{aifit} focuses on video data, and learns to produce feedbacks out of template sentences, based on the comparison between a trainee's and a trainer's motion extracted features. More recently, PoseFix~\cite{delmas2023posefix} adapted the automatic captioning pipeline from \cite{posescript} to create synthetic instructions for a pair of 3D poses sampled from AMASS~\cite{amass}. Those prove useful for pretraining, before finetuning on a small set of human-written texts. The proposed text generation model is a simple text decoder conditioned on pose pairs via cross-attentions. However, it is limited to parameterized pose input (\ie 3D joint rotations), and thus cannot be directly applied to real-world scenarios, as in a fitness coaching application receiving camera input.
In this work, we use our 3D-, visual- and semantic-aware embedding to scale the task on direct image input, without having to train the model on a dataset of real images and textual instructions together.
\section{The PoseEmbroider framework}
\label{sec:main_method}

\begin{figure}[t!]
    \centering
    \includegraphics[width=\textwidth]{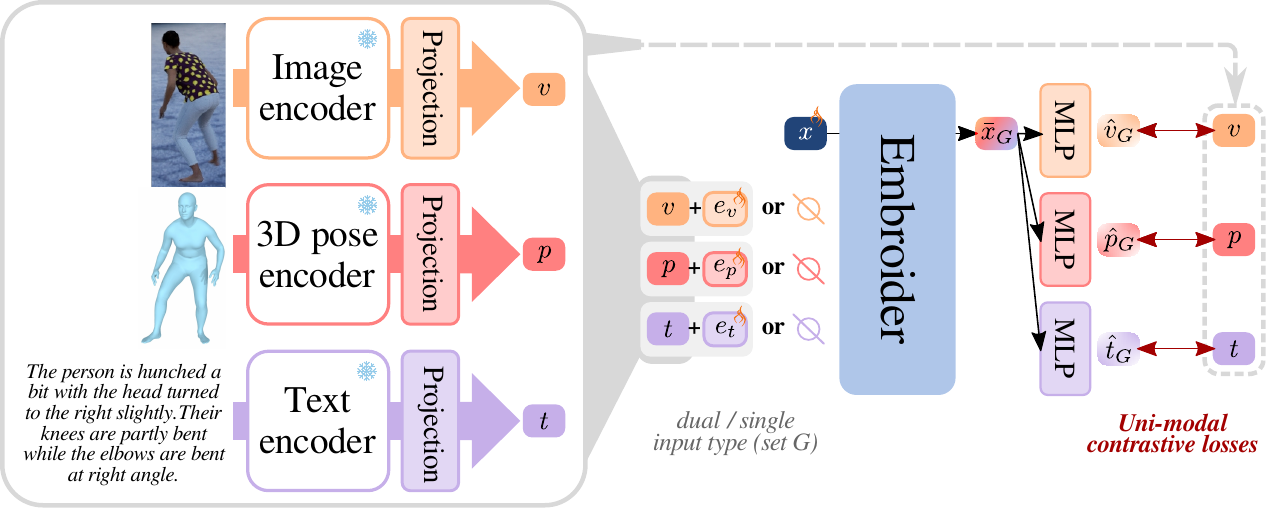} \\[-0.3cm]
    \caption{\textbf{The PoseEmbroider framework.} Each modality is encoded independently by an encoder (left). The PoseEmbroider (right) is a transformer-based model, taking a varying set of modality inputs. It produces a visual-, 3D-, semantic-aware pose representation $\bar{x}$, by embroidering together available inputs. The model is trained using uni-modal contrastive losses between the modality-specific reprojections $\hat{m} \in \{\hat{v}, \hat{p}, \hat{t}\}$ of $\bar{x}$ and the original modality encodings $m \in \{v,p,t\}$. The total objective function accounts for various $\bar{x}_G$, obtained from the set $G$ of input modalities. $x$ and $e_m$ are learnable tokens, `+' denotes an addition.
    }
    \label{fig:embroider}
\end{figure}

We now describe our proposed framework for learning multi-modal enhanced pose representations, see Figure~\ref{fig:embroider} for an illustration. 
Note that the overall design does not rely on specific types or numbers of modalities, allowing for its extension to other domains and sets of modalities.
In this paper, we focus on three modalities: images of people, 3D human poses (parameterized by the rotations of the main SMPL~\cite{smpl} body joints) and text, in the form of fine-grained pose descriptions in natural language. Each of them provide different kinds of information, be it visual, spatial and kinematic, or semantic. We aim to leverage their partial representation of the same abstract concept of human pose, to build a richer pose embedding. 
For simplicity, we assume in what follows that we have a tri-modal dataset, \ie with samples from all modalities for each example.

\subsection{Method}
\label{sec:embroider_method_subsection}

\myparagraph{Encoders.} Similar to other multi-modal methods~\cite{girdhar2023imagebind}, we resort to pretrained uni-modal encoders for each modality. Specifically, we use a Vision Transformer~\cite{dosovitskiy2020image} tuned on human data~\cite{smplx} to encode images; a variant of the VPoser~\cite{smplx} encoder for poses (trained on the main 22 body joints); and a text transformer~\cite{transformer} mounted on top of DistilBERT~\cite{sanh2019distilbert} frozen word embeddings, obtained from a text-to-pose retrieval model~\cite{posescript}. All encoders are used frozen.

\myparagraph{General framework.}
Each modality input is first processed by its respective frozen pretrained encoder, then fed to a modality-specific learnable linear layer followed by a ReLU activation, to select pose-related features and filter out irrelevant details (\eg background information in images). Average pooling further reduces multi-token representations into single vectors (image case). Let $v$, $p$ and $t$ in $\mathbb{R}^d$ denote the corresponding outputs for the image, pose and text of a data triplet respectively. In what follows, we use $m$ to refer to any of them: $m \in M \coloneqq \{v, p, t\}$.

The \emph{PoseEmbroider} mainly consists of a transformer~\cite{transformer}. It takes a variable set of input modalities $G \in S$, in addition of a learnable global token $x$, that will collect and aggregate pose knowledge across all input modalities through the attention mechanisms of the transformer.
We consider any combination of one (\textit{``single input''} type) or two (\textit{``dual input''} type) input modalities, \ie $S \coloneqq \{\{v\}, \{p\}, \{t\}, \{v,p\}, \{v,t\}, \{p,t\}\}$.

Hence, the PoseEmbroider is provided with $\{x\} \cup G$, where a modality-specific learnable token $e_m \in \mathbb{R}^d$ has been added to each input modality encoding in order to inform the transformer about their nature. $e_m$ can be compared to a special kind of learnable positional encoding.

The PoseEmbroider  outputs $\vert G \vert +1$ tokens. Yet, we only consider the first one, noted $\bar{x}_G$, which derives directly from the token $x$ and holds specific information from $G$. It represents the richer, multi-modal informed pose embedding, illustrated as the 3D object in Fig.~\ref{fig:motivation_illustration}. It can be obtained from any set of input modalities, and be used as main pose representation in downstream tasks.

\myparagraph{Training.} To ensure $\bar{x}_G$ carries important visual, spatial \& kinematic and semantic pose information, we compare it to each of the original unimodal encodings. However, we do not perform a direct comparison of $\bar{x}_G$ with each modality encoding $m \in M$, as it would compel all modalities to live in the same space, and eventually lead to the collapse of $\bar{x}_G$ to a representation of \textit{common} information between modalities. Instead, we want $\bar{x}_G$ to be an \textit{enhancement} of its components. Even more, we want it to form sensible postulates for the modalities that did not directly contributed to its derivation.

Thus, to train $\bar{x}_G$, we project it ``back'' to each modality space thanks to expendable modality-specific multi-layer perceptrons~\cite{haykin1994neural_mlp} (MLPs). These yield $\hat{m}_G \in \hat{M}_G \coloneqq \{\hat{v}_G,\hat{p}_G,\hat{t}_G\}$. For a given batch of $B$ training samples, we then compute the uni-modal contrastive loss for each modality $m$, following the widely used InfoNCE~\cite{oord2018representation_infoNCE}:
\begin{equation}
    \mathcal{L}_c(y,z) = - \frac{1}{B} \sum_{i=1}^B \text{log} \frac{\text{exp} \big( \gamma~ \sigma(y_i, z_i) \big) }{\sum_j \text{exp} \big( \gamma~ \sigma(y_i, z_j) \big) },
\end{equation}
where $\gamma$ is a learnable temperature parameter and $\sigma$ is the cosine similarity function defined as $\sigma(y,z) = y^\top z / (\Vert y \Vert \Vert z \Vert)$.
Denoting $\mathcal{M}_G \coloneqq \{(m, \hat{m}_G) \mid m \in M, \hat{m}_G \in \hat{M}_G \text{~of the same modality} \}$, the total loss is then:
\begin{equation}
    \mathcal{L} = \sum_{G \in S} \sum_{\mathcal{M}_G} \mathcal{L}_c(m, \hat{m}_G).
\end{equation}

Metaphorically, if we refer to Figure~\ref{fig:motivation_illustration}, we used available shadows ($G$) to try to infer the 3D object ($\bar{x}_G$), thanks to the PoseEmbroider.
To optimize the latter, we light the object under different angles to check shadow consistency in a soft way ($\mathcal{L}$). Specifically, we do not require the shadows to perfectly match (as it would be the case with a reconstruction loss): we only enforce the ranking of the real object's shadow to be better than another object's shadow.
Actually, during this ``validation'' step, we assume access to all ground-truth shadows: even if one or more modalities were missing from the input, as \eg with $\{p\}$, the loss is applied on \textit{all} available modalities. This design forces $\bar{x}_G$ to be multi-modal aware, beyond being simply multi-modal informed. In other words, the PoseEmbroider aims at providing a strong representation of any (partial) combination of the modalities.

\subsection{Dataset: BEDLAM-Script}
\label{sec:bedlamscript}

A multi-modal model requires multi-modal data. However, there is no existing dataset that gathers images, 3D poses and texts all at once. In fact, perfect 3D pose labeling generally requires expensive and non-scalable in-studio capture. Therefore, datasets with real-world images and good-quality 3D human pose annotations are rare, which motivates the creation of synthetic datasets. 
BEDLAM~\cite{Black_CVPR_2023_bedlam} is the most recent endeavor in this regard. 
It provides rendered sequences of clothed humans in different environments, performing a wide variety of motions extracted from the AMASS dataset~\cite{amass}. It thus comes with ground-truth 3D pose annotations. Previous works~\cite{cai2024smpler} have shown that training on BEDLAM brought the best result for pose estimation over various real data benchmarks~\cite{vonMarcard20183DPW, lin2023osx_ubody, zhang2022egobody}, compared to training on other (including real) datasets.

We thus opt for this dataset. Similar to~\cite{posescript}, we first select a set of $N$ diverse poses by farthest point sampling, \ie sampling iteratively the pose that has the largest mean-per-joint distance with respect to the set of poses already selected. This process makes it possible to efficiently reduce the size of the training ($N$=50k) and validation ($N$=10k) sets while preserving data diversity. We augment each image-pose pair with 3 detailed pose descriptions using the automatic captioning pipeline from PoseScript~\cite{posescript}. Specifically, given 3D joint coordinates, they compute a collection of ``posecodes'' informing about atomic pose configurations (\eg bending of a body part, relative body part positioning, \etc). Those are further converted to natural language description thanks to a set of syntactic rules, merging posecodes that carry similar semantic information. We improve this pipeline to account for head rotations and self-contacts, so as to get better pose descriptions. We do so using a mesh rendering of the pose, and a self-contact detection algorithm~\cite{Mueller_selfcontact_2021} coupled with a semantic segmentation of body vertices. We refer to the resulting dataset of images, 3D poses and text descriptions as BEDLAM-Script, and train our PoseEmbroider framework on it.
While its training involves exclusively synthetic data, we show that the PoseEmbroider produces convincing results on real-world images and human-written texts.

\myparagraph{Data processing.} We consider normalized 3D body poses, \ie with the global rotation set such that the hips are aligned and always facing in the same direction. The motivation is to force the model to extract more general, world-anchored pose knowledge, in contrast to camera-dependent pose information.
While BEDLAM annotations are in SMPL-X~\cite{smplx} format, \ie they include hands, we restrict the 3D pose representation to the main 22 joints of the body. Future work could additionally consider the hands, by also adapting the automatic captioning pipeline to provide such information, \eg as in \cite{lin2024motionx}.
\section{Results on Multi-Modal Retrieval}
\label{sec:multimodal_retrieval}

As a direct side product of its training, the PoseEmbroider framework exhibits multi-modal retrieval abilities. In this section, we report results for any-to-any multi-modal retrieval, and use this task for our ablations.
We additionally showcase qualitative results for edited-retrieval in a multi-modal setting.

\myparagraph{Evaluation metrics.} We consider all possible any-to-any multi-modal retrieval sub-tasks.
This results in 6 single-query and 3 dual-query tasks. 
The standard metric for retrieval evaluation is the recall@K ($R@K$), \ie the percentage of queries whose annotated target appears within the top-$K$ of retrieved elements. We report the average recall over $K \in \{1, 5, 10\}$.

\begin{table*}[t]
    \centering    
    \caption{\textbf{Multi-modal retrieval results.} Models are trained on BEDLAM-Script and evaluated on its validation set. The \textit{total} mRecall is the average of \textit{single} and \textit{dual}, corresponding to the average over all single- and dual-query retrieval tasks respectively. V, P, and T refer to the ``visual'' (image), ``pose'' and ``text'' modalities respectively. The aligner trained on single-input only (first row) corresponds to the idea of ~\cite{girdhar2023imagebind,posescript}.}
    \vspace{-0.3cm}
    \resizebox{\textwidth}{!}{%
    \begin{tabular}{l@{~~}c@{~~}c@{~~}c@{~~~~}c@{~~}c@{~~}c@{~~}c@{~~}c@{~~}c@{~~}c@{~~}c@{~~}c}
    \toprule
    & \multicolumn{3}{c}{mRecall\color{OliveGreen}{$\uparrow$} } & \multicolumn{6}{c}{Single query} & \multicolumn{3}{c}{Dual query} \\
    \cmidrule(lr){2-4} \cmidrule(lr){5-10} \cmidrule(lr){11-13}
    & \textit{total} & \textit{single} & \textit{dual} & \ret{V}{P} & \ret{V}{T} & \ret{P}{V} & \ret{P}{T} & \ret{T}{V} & \ret{T}{P} & \ret{VP}{T} & \ret{PT}{V} & \ret{VT}{P} \\
    \midrule
    \rowcolor{myrowcolor}
    \multicolumn{13}{l}{\textbf{\textit{Representation \& training input subsets}}} \\
    ~Aligner (single-input only) & 72.4 & 66.5 & 78.3 & 77.5 & 46.3 & 75.8 & 76.0 & \Fbest{46.2} & \Sbest{77.5} & 71.3 & 70.7 & 92.8 \\ 
    ~Aligner (dual-input extension) & \Sbest{72.5} & 66.4 & 78.5 & 76.9 & 45.8 & 75.8 & 76.3 & \Sbest{45.9} & \Fbest{77.8} & 72.0 & 70.4 & \Sbest{93.1} \\ 
    \midrule
    ~PoseEmbroider (single input only) & 69.7 & \Sbest{66.7} & 72.7 & \Fbest{80.2} & \Fbest{48.0} & 74.6 & 77.7 & 43.6 & 76.4 & 67.5 & 61.9 & 88.7 \\ 
    ~PoseEmbroider (dual input only) & 71.1 & 58.7 & \Fbest{83.6} & 69.7 & 30.0 & \Fbest{78.0} & \Fbest{78.7} & 26.5 & 69.2 & \Fbest{79.6} & \Fbest{78.2} & 93.0 \\ 
    ~\bf{PoseEmbroider} ($S$) & \Fbest{74.6} & \Fbest{66.9} & \Sbest{82.2} & \Sbest{79.7} & \Sbest{47.8} & \Sbest{76.2} & \Fbest{78.7} & 43.2 & 75.8 & \Sbest{77.9} & \Sbest{75.1} & \Fbest{93.7} \\ 
    \midrule
    \rowcolor{myrowcolor}
    \multicolumn{13}{l}{\textbf{\textit{PoseEmbroider architecture}}} \\
    ~MLP core & 73.4 & \Sbest{66.5} & 80.3 & \Sbest{79.9} & 46.4 & 75.5 & \Fbest{79.2} & 41.9 & \Fbest{76.2} & \Sbest{76.8} & 71.1 & \Sbest{93.0} \\ 
    ~Trans.\ core, no projection heads & \Sbest{73.5} & 66.4 & \Sbest{80.5} & \Fbest{80.2} & \Sbest{46.7} & \Sbest{75.9} & 76.5 & \Fbest{43.6} & 75.6 & 75.0 & \Sbest{73.6} & 93.0 \\ 
    ~Transf.\ core, w/ projection heads (\textbf{proposed}) & \Fbest{74.6} & \Fbest{66.9} & \Fbest{82.2} & 79.7 & \Fbest{47.8} & \Fbest{76.2} & \Sbest{78.7} & \Sbest{43.2} & \Sbest{75.8} & \Fbest{77.9} & \Fbest{75.1} & \Fbest{93.7} \\ 
    \bottomrule
    \end{tabular}
    }
    \label{tab:embroider_table}
\end{table*}

\myparagraph{Alignment baselines.}
To highlight the benefits of the PoseEmbroider representation over a more typical alignment-based representation, we further introduce the Aligner model (similar in number of learnable parameters to the full PoseEmbroider).
Unlike the PoseEmbroider, the frozen pretrained uni-modal encoders in the Aligner are followed by \textit{deep} learnable modality-specific projection heads (\ie MLPs, as opposed to single modality-specific layers leading to a shared transformer). The MLP heads are trained with pair-wise and triplet-wise alignment losses to produce a joint embedding space:

\begin{equation}
    \mathcal{L} = \sum_{G \in S} \sum_{m \notin G} \mathcal{L}_c \Big(m, \frac{1}{\vert G \vert} \sum_{q \in G} q \Big).
\end{equation}

Simply put, there is one contrastive loss term $\mathcal{L}_c(m_1,m_2)$ for each pair of modalities (single input), and one for each kind of dual input, where the dual query representation is computed as the average of its components' features. We denote as \textit{Aligner (single-input only)} the model trained solely on $S' \coloneqq \{\{v\}, \{p\}, \{t\}\}$, and as \textit{Aligner (dual-input extension)} the model trained on $S$.

The \textit{Aligner (single-input only)} can be thought as a version of ImageBind~\cite{girdhar2023imagebind} applied to the human pose domain, or as a version of the PoseScript retrieval model~\cite{posescript} connected with an image network. Yet, different from these approaches, and \textit{to allow a fair comparison} with the PoseEmbroider, the core encoders are not optimized: solely the MLP heads are (and are trained on BEDLAM-Script as well). Eventually, the \textit{Aligner (dual-input extension)} explicitly integrates compositionality in the training objective, conversely to~\cite{girdhar2023imagebind}.

\myparagraph{Quantitative results} are presented in Table~\ref{tab:embroider_table}.
First of all, they reveal that our proposed PoseEmbroider (row 5) outperforms the best Aligner baseline (row 2) by 2.9\% (mRecall), showing particular progress with respect to dual queries (4.7\%). It suggests that the PoseEmbroider not only enhances single-modality encodings but also effectively combines their knowledge. It is especially blatant for pose retrieval, where the use of both image and text as input improves over using each alone (+17.6\% and +23.6\%, respectively). Other cases (\eg image retrieval) hint at the PoseEmbroider ability to extract intel from the most informative modality.
Note that results involving both the visual and textual modalities are the lowest for all models because they are the most ambiguous (occlusions, truncations by image boundaries, incomplete/imprecise descriptions).

\myparagraph{Query set ablation.}
It stems that the Aligner design is not very sensitive to an enhanced optimization using input combinations (row 2 \vs row 1). In contrast, it appears clearly that training on various sets of inputs ($S$) is valuable to the PoseEmbroider, as it improves the mean Recall by +7\% compared to using single inputs only (row 5 \vs row 3). Unsurprisingly, we also observe better performance for single queries when considering single input types during training (+13.6\%), and likewise for dual queries (+15.0\%, row 3 \vs row 4). Eventually, the best performance for all retrieval tasks is reached when using both query types.

\myparagraph{Architecture ablation.} To justify the PoseEmbroider design, we first replace the transformer with an MLP (row 6 in Table~\ref{tab:embroider_table}). In this setting, all modality encodings at stake are added together and fed to an MLP, whose output plays the role of $\bar{x}_G$. Unlike concatenation, the addition operation allows the model to be run on various subsets of modalities.
This model is trained with the same objective as the proposed PoseEmbroider. Next, we ablate the re-projection heads which make it possible to use uni-modal contrastive losses. This configuration has a training objective similar to a regular alignment model.
Results reveal that the transformer version is slightly more powerful than the MLP version (+1.6\%), and that the re-projection heads are valuable (+1.5\%), especially in the dual-query case (+2.1\%). 

\begin{figure}[t]
    \centering
    \includegraphics[width=\textwidth]{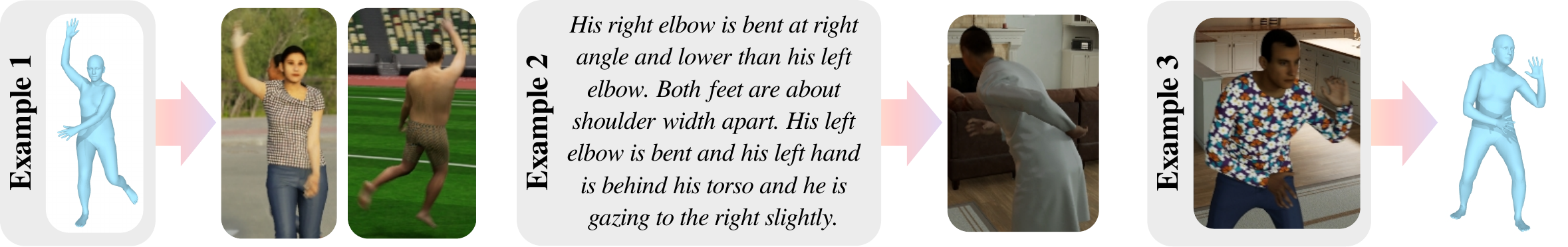} \\[-0.3cm]
    \caption{\textbf{Qualitative examples of any-to-any multi-modal retrieval} on the validation split of BEDLAM-Script, for diverse input and output modalities.}
    \label{fig:qual_a2a_mm_retrieval}
    \vspace{-0.3cm}
\end{figure}

\myparagraph{Qualitative results.}
Figure~\ref{fig:qual_a2a_mm_retrieval} presents some results for any-to-any multi-modal retrieval, demonstrating that the PoseEmbroider efficiently associates the different pose modalities and exhibits human pose understanding.

\begin{figure}[t!]
    \centering
    \includegraphics[width=\textwidth]{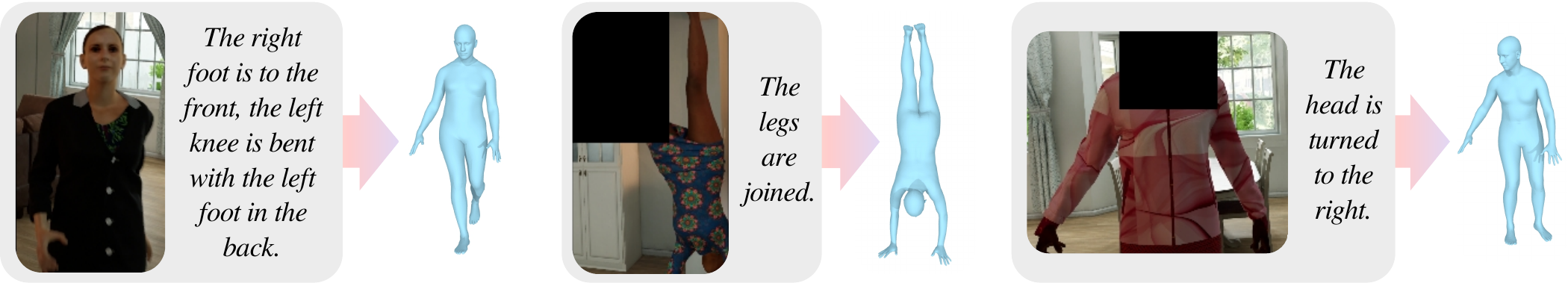} \\[-0.3cm]
    \caption{\textbf{Qualitative examples of edited-retrieval in a multi-modal setting} on BEDLAM-Script. Texts specify new traits with respect to the original pose shown in the image. Artificial occlusion is created by overlaying a black rectangle on the image.}
    \label{fig:qual_edited_retrieval}
    \vspace{-0.3cm}
\end{figure}

We further consider the special use case of ``edited-retrieval'' where, for instance, the user is looking for a 3D pose similar to the one depicted in an image, yet a little different. The parts of the image showing unwanted pose traits are masked while supplementary information is provided through text input. We observe that the model is able to leverage and combine information from both modalities to find relevant poses. Examples are shown in Figure~\ref{fig:qual_edited_retrieval}.

\section{Results on human pose instruction generation}
\label{sec:pose_instruct}

Human pose instruction generation~\cite{fixmypose,delmas2023posefix,aifit} consists in generating directions in natural language to correct a human pose. 
This task has direct application in at-home fitness coaching, to provide automatic feedback.
It can be solved with a text decoder, conditioned on both the \textit{source} pose $A$ (the trainee's) and the \textit{target} pose $B$ (the trainer's).
The poses could be highly similar, and differ only in subtle aspects. Hence, this task typically requires a fine-grained semantic understanding of the human pose. 
Previous works~\cite{delmas2023posefix} have proposed methods that operate on 3D pose inputs, however they cannot handle real-world scenarios where the user simply works out in front of their phone camera.
To further evaluate our proposed PoseEmbroider, we replace the original 3D pose encoder in \cite{delmas2023posefix} by our pretrained PoseEmbroider. This configuration makes it possible to train the text generation model on reliable 3D poses, and seamlessly transfer to visual inputs, without requiring further training. 

\myparagraph{Datasets: BEDLAM-Fix, PoseFix-OOS.}
Available datasets for this task include PoseFix~\cite{delmas2023posefix} and FixMyPose~\cite{fixmypose}. Both have approximately the same training size (4-6k), and provide human-written annotations, however the first one pairs 3D human poses from a wide variety of AMASS~\cite{amass} sequences, while the second one links highly synthetic images of poses extracted from about 20 Unity dance motions. We resort to PoseFix for finetuning the text decoder, and restrict to out-of-sequence (OOS) pairs to eliminate noise stemming from global rotation changes.
Similar to BELDAM-Script (Section~\ref{sec:bedlamscript}), we process BEDLAM~\cite{Black_CVPR_2023_bedlam} to create pretraining data, following the same procedure as in~\cite{delmas2023posefix}. Specifically, we sample pose pairs from BEDLAM-Script by enforcing both semantic similarity and minimal pose difference constraints. We consider both pairs of poses performed by the same subject (\ie with the same appearance, environment and motion) and different subjects. We further run their automatic \textit{comparative} pipeline on the 3D poses to obtain synthetic instruction texts.
We end up with 54k (resp. 12k) training (resp. validation) pairs.

\begin{figure}[t!]
    \centering
    \includegraphics[width=\textwidth]{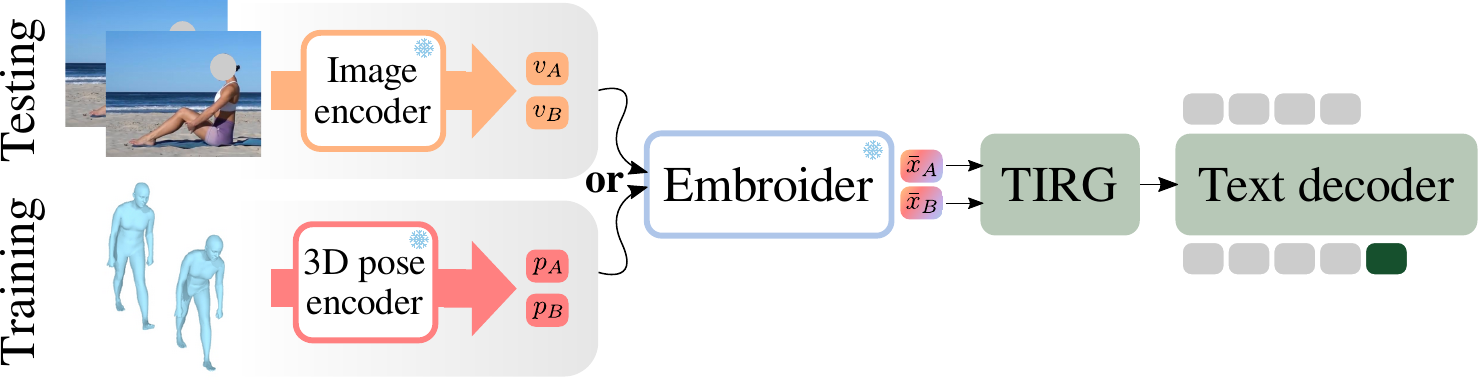} \\[-0.3cm]
    \caption{\textbf{The pose instruction generation model.} We train the model on pairs of poses $(p_A,p_B)$ and use our frozen PoseEmbroider to encode them. These two embeddings are fused with TIRG~\cite{vo2019composing}, whose output is used to condition an auto-regressive transformer text decoder via cross-attentions. At test time, the trained model can be directly applied on poses, images or a mix of both.}
    \label{fig:txt_generation_methods}
\end{figure}

\myparagraph{Method overview.}
We use a similar method as~\cite{delmas2023posefix}: the elements A and B are encoded through siamese networks, and fused thanks to the TIRG module~\cite{vo2019composing} to condition an auto-regressive text decoder.
We experiment with both the \textit{Aligner (dual-input extension)} and the PoseEmbroider for encoding the inputs. To train, we use their cached (frozen) pretrained features, obtained from the 3D poses data. Sole the fusing module and the text decoder are being learned. At inference time, we use any combination of 3D poses and visual input, see Figure~\ref{fig:txt_generation_methods}.

\myparagraph{Baseline.}
We use our best Aligner to represent the PoseFix baseline~\cite{delmas2023posefix}. Indeed, their encoder takes only 3D pose input, and is trained specifically for the task, alongside the text decoder. Yet, we aim to compare off-the-shelf representations, and further empower inference from visual input.

\myparagraph{Evaluation metrics.} Following~\cite{delmas2023posefix}, we retrain an instruction-to-pair retrieval model on the aforementioned datasets to assess the semantic content of the generated text through R-precision (with a larger, harder pool of 200). This complements the typical n-gram based NLP metrics (BLEU-4~\cite{Papineni02bleu}, Rouge-L~\cite{lin2004rouge} and METEOR~\cite{banarjee2005meteor}) which evaluate formulation similarity with the reference text.

\begin{table*}[t]
    \centering
    \caption{\textbf{Text generation results for different query types.} Models are trained on BEDLAM-Fix using pairs of poses only, and evaluated on the associated validation split for queries of different natures. We further finetune the text decoder on a mix of BEDLAM-Fix and PoseFix-OOS data, and report results on the test set of PoseFix-OOS. The Aligner baseline represents~\cite{delmas2023posefix}.} 
    \vspace{-0.3cm}
    \resizebox{\textwidth}{!}{%
        \begin{tabular}{l@{~~~~}l@{~~~~}c@{~~}c@{~~}c@{~~}c@{~~}c@{~~}c@{~~}}
    \toprule
    \multirow{2}{*}{Dataset (query type)} &
    \multirow{2}{*}{Representation} & \multicolumn{3}{c}{R Precision\color{OliveGreen}{$\uparrow$} } & \multicolumn{3}{c}{NLP\color{OliveGreen}{$\uparrow$}} \\
    \cmidrule(lr){3-5} \cmidrule(lr){6-8}
    & & R@1 & R@2 & R@3 & BLEU-4 & ROUGE-L & METEOR \\
    \midrule
    \multirow{2}{*}{BEDLAM-Fix ($P_A$, $P_B$)} &
    Aligner
    & 31.7 & 42.4 & 49.4 & 32.9 & 42.2 & 40.1 \\ 
    & \textbf{PoseEmbroider} & \Fbest{43.1} & \Fbest{55.1} & \Fbest{62.0} & \Fbest{33.0} & \Fbest{42.6} & \Fbest{40.7} \\ 
    \midrule
    \multirow{2}{*}{BEDLAM-Fix ($V_A$, $V_B$)} &
    Aligner
    & 12.3 & 19.2 & 23.7 & \Fbest{27.8} & \Fbest{37.8} & \Fbest{38.9} \\ 
    & \textbf{PoseEmbroider} & \Fbest{15.6} & \Fbest{22.6} & \Fbest{27.1} & 27.4 & 37.2 & 37.3 \\ 
    \midrule
    \multirow{2}{*}{BEDLAM-Fix ($P_A$, $V_B$)} &
    Aligner
    & 16.6 & 24.2 & 29.8 & 28.9 & \Fbest{38.7} & \Fbest{40.4} \\ 
    & \textbf{PoseEmbroider} & \Fbest{21.2} & \Fbest{29.6} & \Fbest{35.6} & \Fbest{29.8} & 38.3 & 38.6 \\ 
    \midrule
    \multirow{2}{*}{BEDLAM-Fix ($V_A$, $P_B$)} &
    Aligner
    & 21.0 & 29.2 & 34.9 & 30.3 & 39.1 & 39.7 \\ 
    & \textbf{PoseEmbroider} & \Fbest{29.2} & \Fbest{38.3} & \Fbest{44.2} & \Fbest{30.6} & \Fbest{39.3} & \Fbest{40.4} \\ 
    \midrule
    \multirow{2}{*}{BEDLAM-Fix ($V_A$+$P_A$, $V_B$+$P_B$)} &
    Aligner
    & 26.5 & 36.4 & 42.4 & \Fbest{33.1} & 41.0 & \Fbest{40.1} \\ 
    & \textbf{PoseEmbroider} & \Fbest{33.8} & \Fbest{45.0} & \Fbest{51.9} & 31.3 & \Fbest{41.3} & 39.5 \\ 
    \midrule
    \multirow{2}{*}{PoseFix-OOS ($P_A$, $P_B$)} &
    Aligner
    & 19.5 & 27.6 & 34.8 & 8.0 & 27.0 & 28.0 \\ 
    & \textbf{PoseEmbroider} & \Fbest{27.4} & \Fbest{36.8} & \Fbest{42.9} & \Fbest{10.2} & \Fbest{28.8} & \Fbest{29.3} \\ 
    \bottomrule
    \end{tabular}
    }
    \label{tab:txt_generation_table}
\end{table*}

\myparagraph{Quantitative results} are reported in Table~\ref{tab:txt_generation_table}, for different sets of inputs, depending on the respective nature of elements $A$ and $B$ at test time (\ie 3D pose or image).
The PoseEmbroider representation outperforms the Aligner representation in all cases, particularly when both inputs are 3D poses (+36\% R@1 on BEDLAM-Fix, +41\% on PoseFix-OOS), despite sharing the same 3D pose encoder at the core. This suggests that the PoseEmbroider is indeed capable of \textit{enhancing} semantic pose representations.
Notably, 3D pose inputs yield better results than visual inputs, which are inherently less reliable (occlusions).
Since instruction generation is driven by element $B$, it makes sense to find better results for the setting $(V_A, P_B)$ than $(P_A, V_B)$, when compared to $(V_A, V_B)$ (+87\% \vs +36\%). The setting $(V_A, P_B)$ typically corresponds to that of a fitness application scenario, involving camera input from the user and clean, 3D pose registrations of the target pose.

\begin{figure}[t!]
    \centering
    \includegraphics[width=\textwidth]{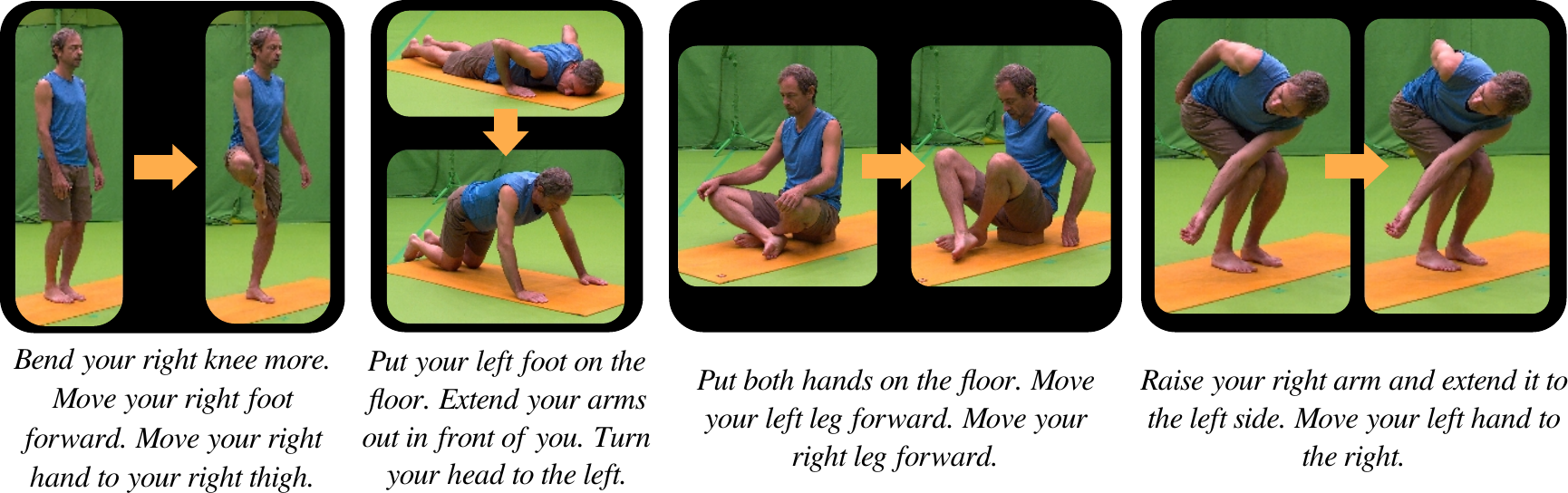} \\[-0.3cm]
    \caption{\textbf{Instruction generations on real-world images using the PoseEmbroider pose representation}. The text model was trained using the frozen PoseEmbroider embeddings of 3D poses only. The generated text is shown below each image pairs.}
    \label{fig:txt_gen_qual}
    \vspace{-0.3cm}
\end{figure}

\myparagraph{Qualitative results.}
In Figure~\ref{fig:txt_gen_qual}, we present examples of generated instructions for real-world input images, illustrating the steps for performing Yoga poses.
While the text generation model was trained using only a dataset of 3D poses and texts, the PoseEmbroider makes it possible to transfer to image input.

\section{Results on SMPL regression}
\label{sec:pose_estimation}

We showcase results for the task of SMPL regression, where the goal is to predict the pose and shape parameters of the SMPL body model~\cite{smpl} for a given input data of any modality. This task is known as 3D Human Mesh Recovery~\cite{Kanazawa_2018_CVPR}, when applied on images.
We proceed similarly as before, and train a neural head to predict SMPL parameters from pretrained, frozen features of the PoseEmbroider and the Aligner, here obtained from image input for training. We use the standard iterative residual network from~\cite{Kanazawa_2018_CVPR} to predict the joint rotations from the mean parameters, and an MLP to regress shape coefficients.

\begin{figure}[t]
    \centering
    \includegraphics[width=\textwidth]{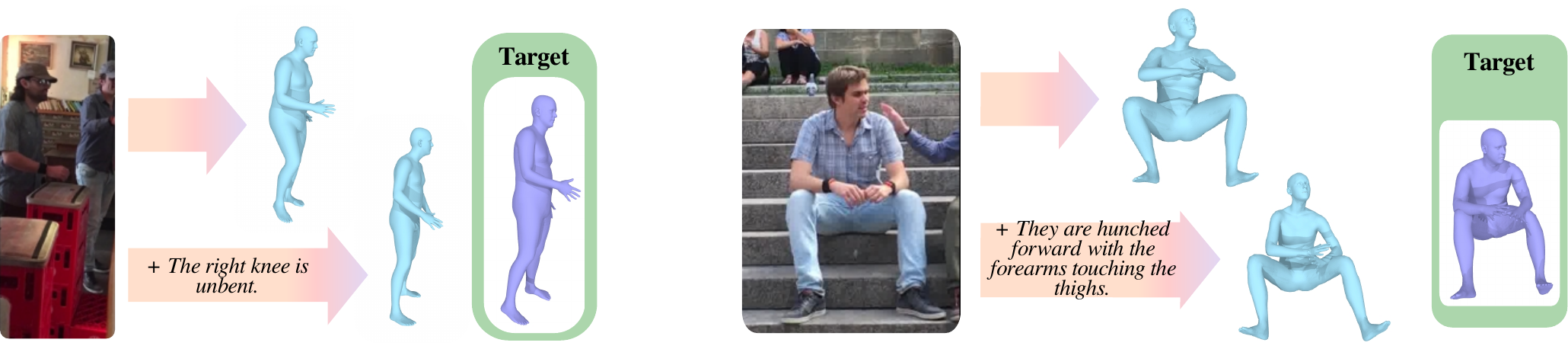} \\[-0.3cm]
    \caption{\textbf{Image-based SMPL regression} with an optional text hint.}
    \label{fig:pose_estimation_qual}
\end{figure}

In Table~\ref{tab:pose_estimation_table}, we report the average pa-MPJPE (Procrustes-aligned Mean Per Joint Position Error) on BEDLAM-Script and 3DPW~\cite{vonMarcard20183DPW}. We note that the state-of-the-art SMPLer-X Huge model~\cite{cai2024smpler} achieves 43 and 41mm for each respectively, and that our PoseEmbroider with the trained SMPL-head is only subpar by 6 and 12mm, while (1) leveraging a smaller ViT (base) for encoding, (2) not training the input representation end-to-end, and (3) training the regression head on 50K synthetic samples only (thus exposing the domain gap on 3DPW).
Interestingly, our model improves by +11\% when provided text cues. The PoseEmbroider design makes it possible to process added textual information without any retraining, \eg to refine estimations as illustrated in Figure~\ref{fig:pose_estimation_qual}.

\begin{table*}[t]
    \centering
    \caption{\textbf{SMPL regression results for different representations and inputs.} The regression head is trained solely on BEDLAM-Script, with frozen image-based features of the Aligner/PoseEmbroider models. We report the pa-MPJPE in mm with the ground truth pose, on BEDLAM-Script (validation set) and 3DPW~\cite{vonMarcard20183DPW} (test set).
    }
    \vspace{-0.3cm}
    \resizebox{0.7\textwidth}{!}{%
    \begin{tabular}{l@{~~~~}c@{~~}c@{~~}}
    \toprule
    pa-MPJPE\color{OliveGreen}{$\downarrow$} & BEDLAM-Script & 3DPW \\
    \midrule
    ~Aligner (image) & 50 & 54 \\
    ~PoseEmbroider (image) & 49 & \Fbest{53} \\
    ~\textbf{PoseEmbroider} (image+text) & \Fbest{44} & - \\
    \bottomrule
    \end{tabular}
    }
    \label{tab:pose_estimation_table}
\end{table*}

\section{Discussion}
\label{sec:dicussion}

\myparagraph{Conclusion.}
We have introduced the PoseEmbroider framework, which derives visual-, 3D-, semantic-aware pose representations. Instead of \textit{aligning} to fit \textit{shared} information across modalities, it is trained to \textit{combine} and thus \textit{enrich} single-modality pose representations. Beyond its direct use in any-to-any multi-modal retrieval, the proposed versatile representation can be leveraged for complex downstream tasks requiring fine-grained human pose understanding, such as pose instruction generation or SMPL regression.

\myparagraph{Limitations and future work.}
Future improvement could come from employing more aggressive losses (\eg attempting to predict target features instead of solely learning to match them), training on more data (the 50k training samples pale in comparison to the 400 millions pairs CLIP was trained on), or incorporating a broader set of modalities (depth maps, 2D keypoints \etc). The described training procedure resorts to a single tri-modal dataset. Yet, we can envision learning from a set of uni-modal and bi-modal datasets, each coming with different groups of modalities.

\small{\myparagraph{Acknowledgments.} This work is supported by the Spanish government with the project MoHuCo PID2020-120049RB-I00, and by NAVER LABS Europe under technology transfer contract ‘Text4Pose’.}


%
%
\bibliographystyle{splncs04}
\bibliography{main}

\clearpage
\section*{Supplementary Material}

\appendix
\setcounter{table}{0}
\renewcommand{\thetable}{A\arabic{table}}
\setcounter{figure}{0}
\renewcommand{\thefigure}{A\arabic{figure}}
\setcounter{equation}{0}
\renewcommand{\theequation}{A\arabic{equation}}

In this supplementary material, we present examples of tri-modal data samples from BEDLAM-Script and BEDLAM-Fix (Section~\ref{app:datasets}), and complete the explanations given in the main paper about these augmentations of BEDLAM (\eg image selection criteria). We provide additional qualitative results and analysis (including limitations) of our models in Section~\ref{app:added_qual}. The original annotations associated to the queries presented in the main paper are available in Section~\ref{app:ground_truth}. Finally, we give implementation details in Section~\ref{app:implem_details} 
and discuss responsibility to human subjects (Section~\ref{app:responsibility}).

\section{More about BEDLAM-Script and BEDLAM-Fix}
\label{app:datasets}

\begin{figure}[ht]
    \centering
    \includegraphics[width=\textwidth]{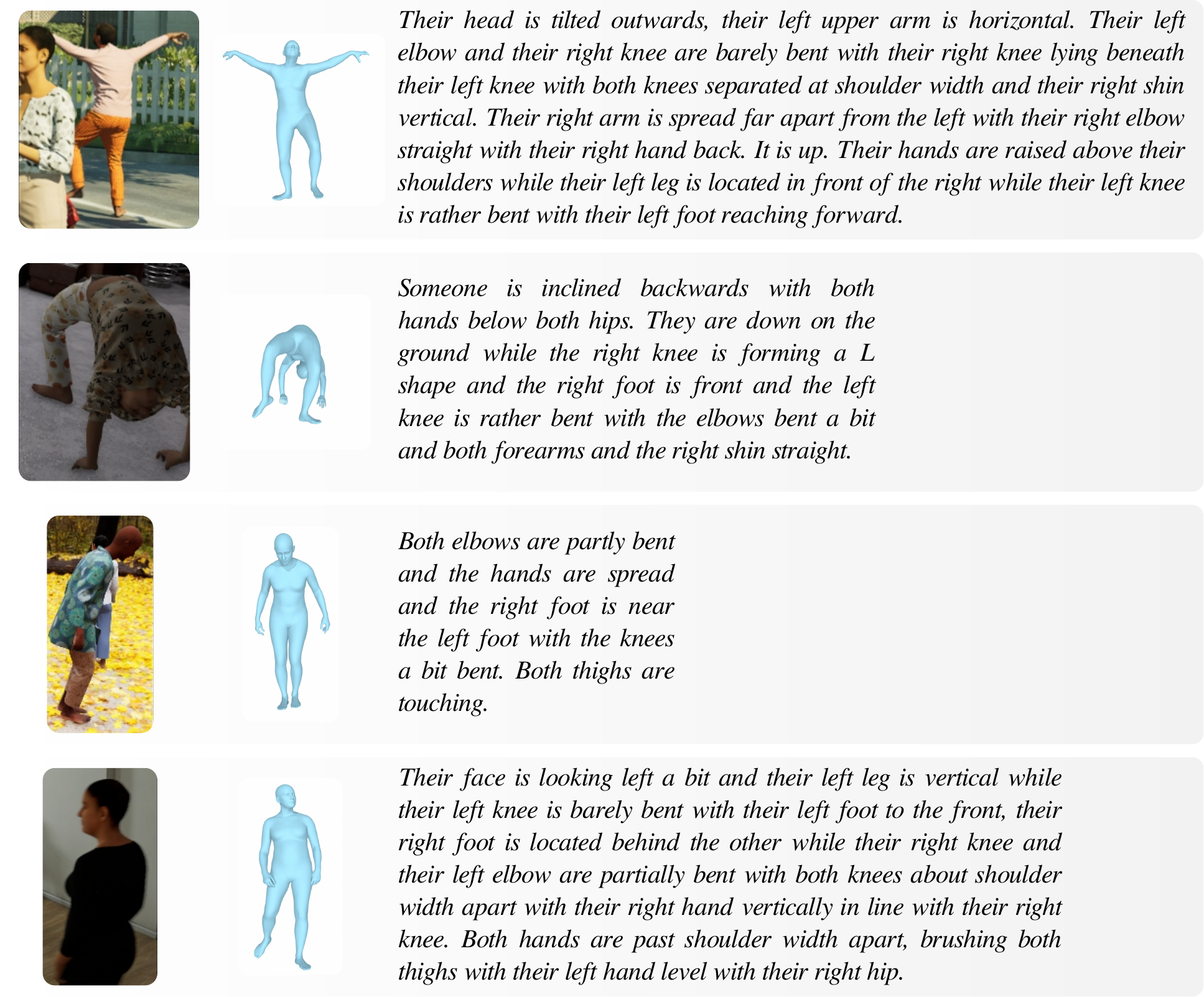} \\[-0.3cm]
    \caption{\textbf{Examples from BEDLAM-Script.} A text describes the 3D pose represented in the image.}
    \label{fig:dataset_examples_script}
    \vspace{-0.3cm}
\end{figure}

\begin{figure}[ht]
    \centering
    \includegraphics[width=\textwidth]{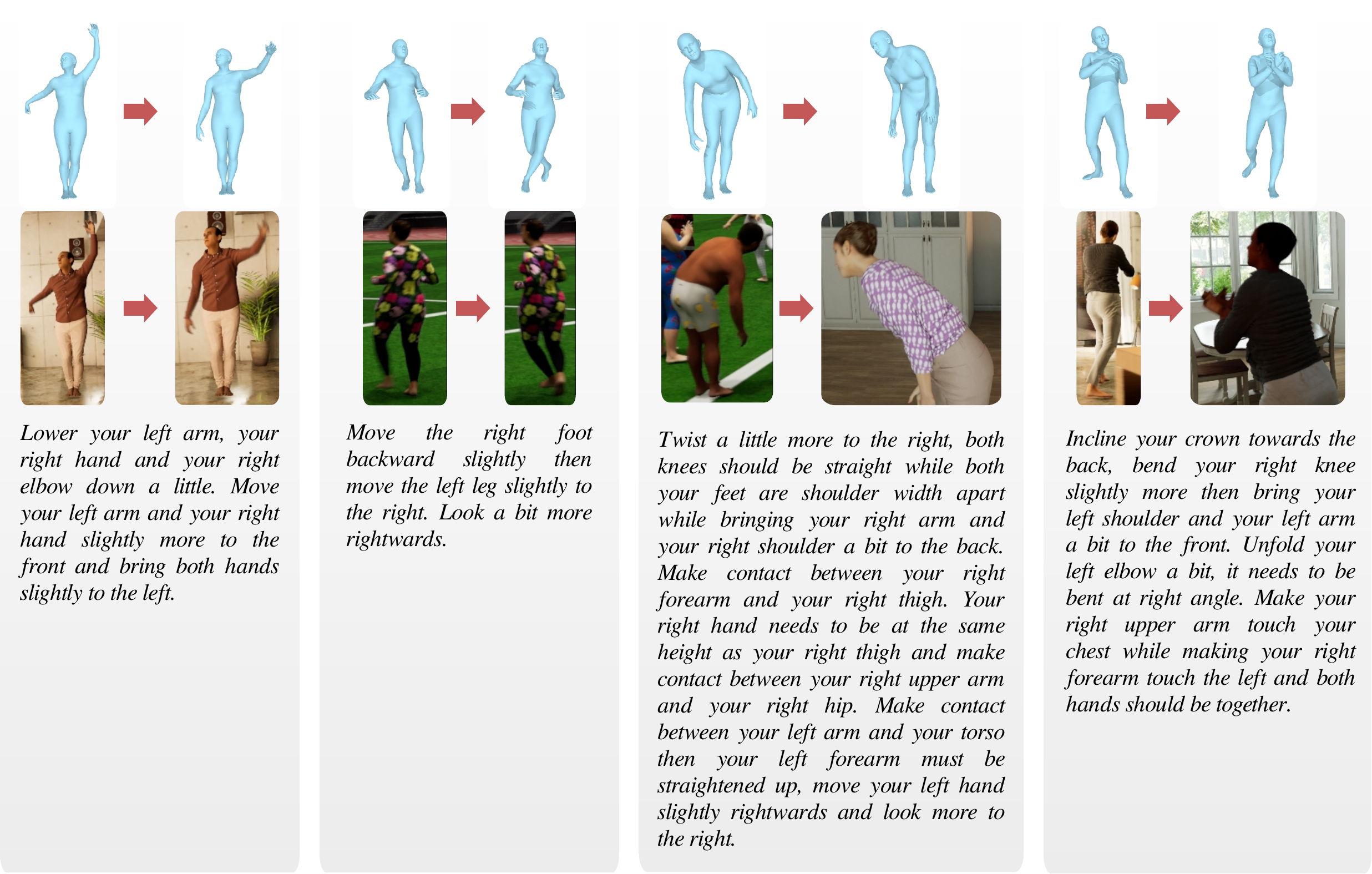} \\[-0.3cm]
    \caption{\textbf{Examples from BEDLAM-Fix.} A textual instruction explains how to go from element A (left side of the arrow) to element B (right side of the arrow). Elements A and B can be either images and/or 3D poses.}
    \label{fig:dataset_examples_fix}
    \vspace{-0.3cm}
\end{figure}

\myparagraph{Dataset examples.}
Figure~\ref{fig:dataset_examples_script} shows some examples of tri-modal samples from BEDLAM-Script. Figure~\ref{fig:dataset_examples_fix} does the same for BEDLAM-Fix.

\myparagraph{Additional details about data selection.}
To select samples from BEDLAM, we begin by a filtering of the images based on joint visibility, then proceed with the farthest pose sampling algorithm described in the main text of this paper. Here is the list of criteria used to select people images:
\begin{itemize}[noitemsep,topsep=0pt,label=$\bullet$]
    \item At least 16 of the main body joints had to be within the image boundaries (although they could be subject to occlusion).
    \item The person was required to be at the forefront (\ie, positioned closest to the front compared to other individuals in the same image). However, we relaxed this condition slightly by also considering individuals positioned further in the background, provided that at least 70\% of their bounding box did not overlap with the bounding box of someone positioned closer to the front.
    \item At least one side of the human bounding box (upscaled by a factor 1.1) had to be more than 224 pixels.
\end{itemize}
Paired elements (poses, images) in BEDLAM-Fix are part of BEDLAM-Script.
\section{Additional qualitative results}
\label{app:added_qual}

\subsection{Text instruction generation}

\begin{figure}[t]
    \centering
    \includegraphics[width=\textwidth]{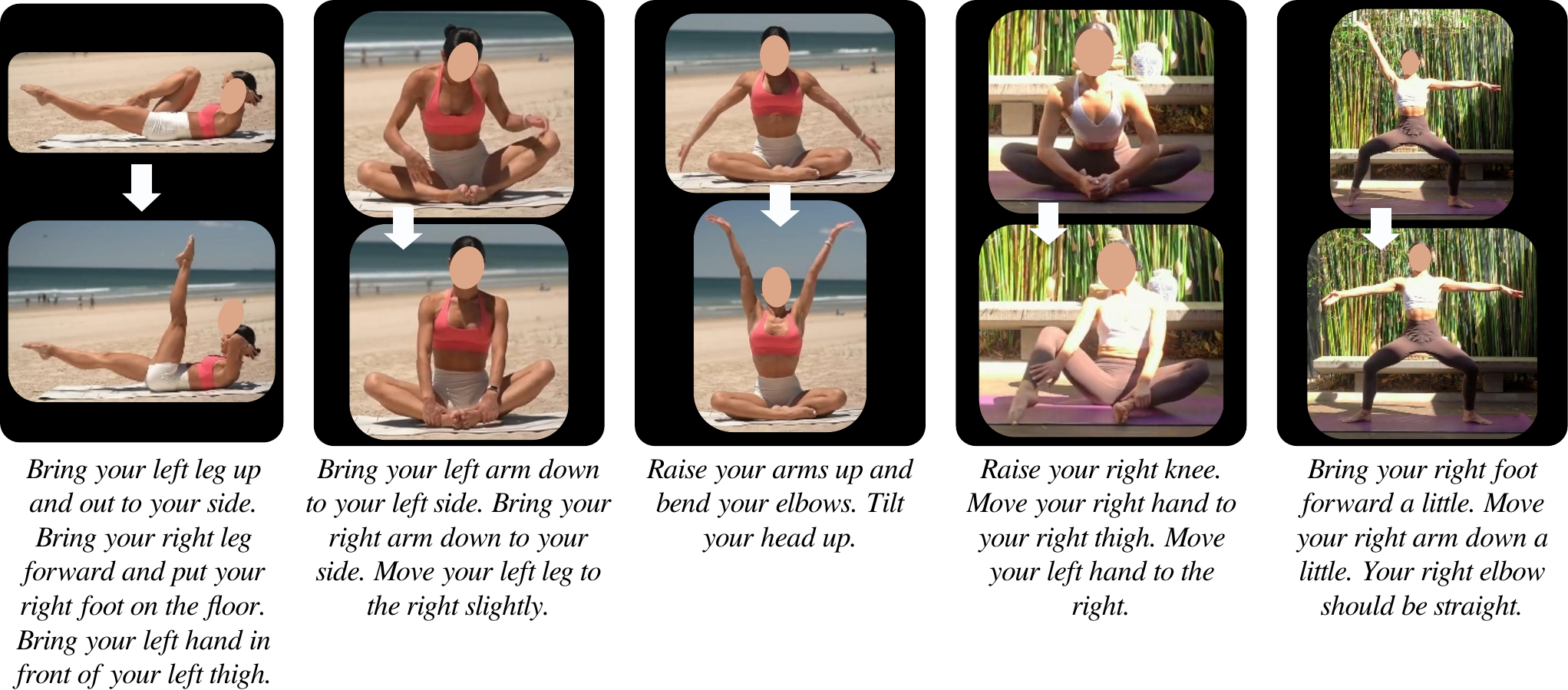} \\[0.1cm]
    \includegraphics[width=\textwidth]{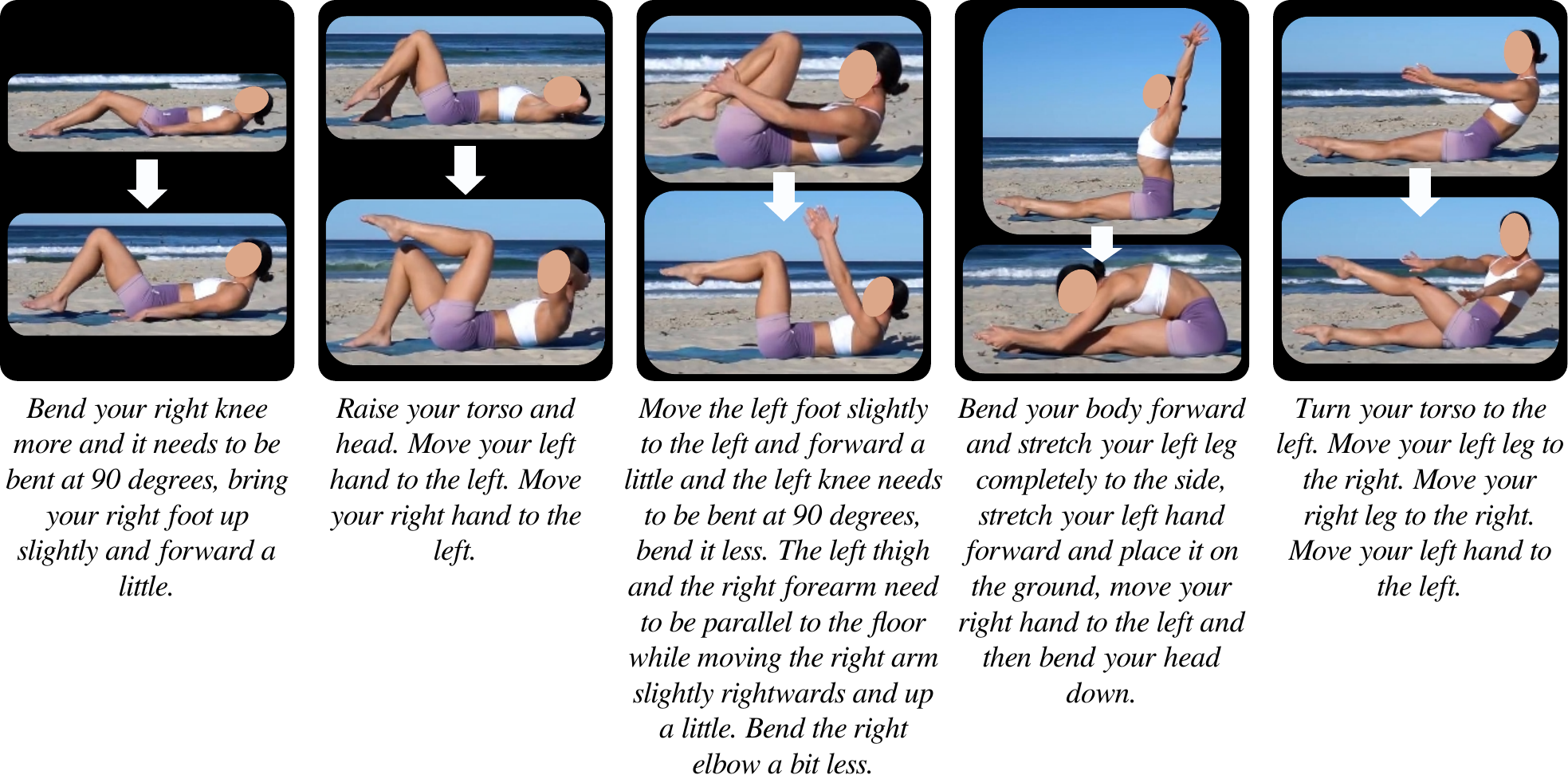} \\[-0.3cm]
    \caption{\textbf{Instruction generations on real-world images using the PoseEmbroider pose representation}. The instruction generation model was trained using the PoseEmbroider representations of 3D poses only. The generated text is shown below each image pairs. We occluded the faces to preserve privacy.}
    \label{fig:txt_gen_qual_nicole}
\end{figure}

The automatic generation of pose instructions has several applications. It makes it possible to give correctional feedback to a trainee by comparing their pose to a trainer's. It also allows automatic narration to accompany sport training videos. In Figure~\ref{fig:txt_gen_qual_nicole}, we provide qualitative results on real-world images depicting pilates moves. These pictures were obtained from YouTube videos of pilates classes.

We notice that the text instructions globally fit the different situations, especially for sitting and standing poses (four last columns of the first row).
In particular, we find that the model is able to distinguish a set of fine-grained body changes, about arm position (1st row, columns 2,3 and 5), leg or arm bending (1st row, column 4; 2nd row, columns 1 and 3), body twist (2nd row, columns 2, 4 and 5), head rotation (1st row, middle example) and so forth.

However, there are still small perception mistakes (\eg \textit{first image of the first row}: the right foot is not exactly on the floor; \textit{middle image of the first row}: the elbows should not really be bent; \textit{fourth image of the first row}: the right hand is rather on the right calf than the right thigh, but the confusion probably comes from the fact that the right wrist appears to be touching the right thigh). One recurrent behavior is the output of instructions like ``move body part X to the right/left'', which do not always seem to really apply, from a human perspective. This likely stems from the fact that the bodies in Figure~\ref{fig:txt_gen_qual_nicole} are only visible from the side -- thus preventing a fair depth estimate.

Globally, the study of several qualitative examples reveals that the model struggles with lying down poses. For instance, in the middle example of the second row, it believes the ground is the left side of the image (hence ``the left thigh and the right forearm need to be \textit{parallel to the floor}''). This problem comes directly from the pose representation, and can be explained by the low number of lying down poses in the training data. The frequency of such poses in the training batches could be increased to mitigate this issue.

Other typical failure cases include when the person is only partially visible (\eg lower body truncated by the image boundaries). In these conditions, the text generation model would often output average instructions about the legs and forget about the main differences of the upper body. One way to alleviate this limitation could be to \textit{also} train the model on truncated images input (instead of 3D poses only), and to consider instructions that only mention differences about the visible body parts. Two main aspects of our method makes this conceivable. First, the PoseEmbroider can similarly treat image and 3D pose input: it provides a modality-agnostic representation to the text decoder. Second, the model can be trained efficiently on synthetic data, using the automatic pipeline from \cite{delmas2023posefix}, which can be modified so as to produce instructions involving a specific set of body joints (\ie, those visible in the images).

\subsection{Any-to-any retrieval}

\begin{figure}[t]
    \centering
    \includegraphics[width=\textwidth]{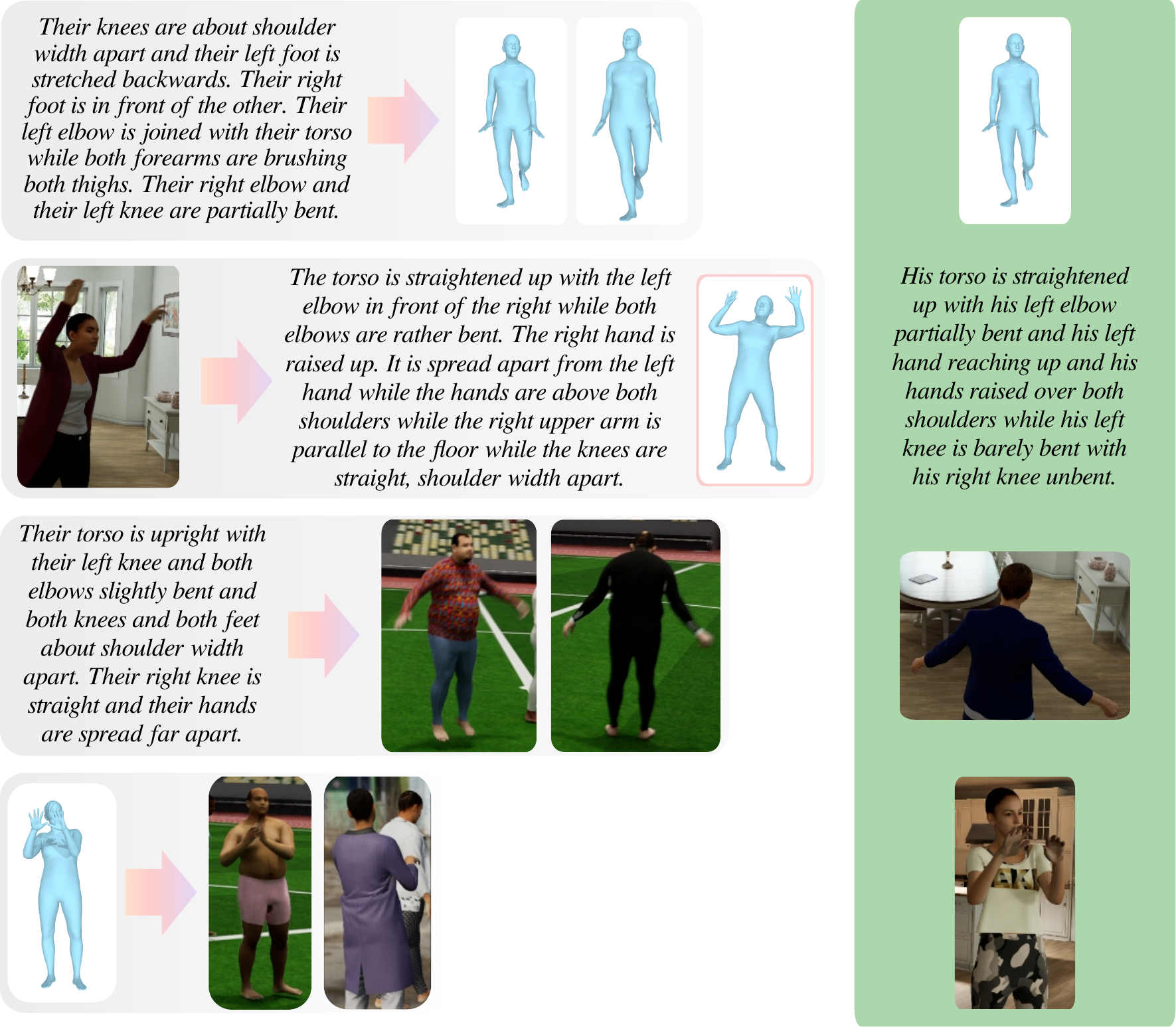} \\[-0.3cm]
    \caption{\textbf{Qualitative examples of any-to-any multi-modal retrieval} on the validation split of BEDLAM-Script. We show either top-1 or top-2 results for several types of single input and output modalities. Original paired modalities for the queries on the left are shown in the green box on the right. To ease reading, we additionally show the 3D pose associated retrieved texts and give it a pink border.
    }
    \label{fig:more_any2any}
\end{figure}

We show a few more examples of any-to-any retrieval in Figure~\ref{fig:more_any2any}. We see that our model produces reasonable results for different types of query and target modalities.

\section{Original annotations for retrieval results}
\label{app:ground_truth}

\begin{figure}[t]
    \centering
    \includegraphics[width=\textwidth]{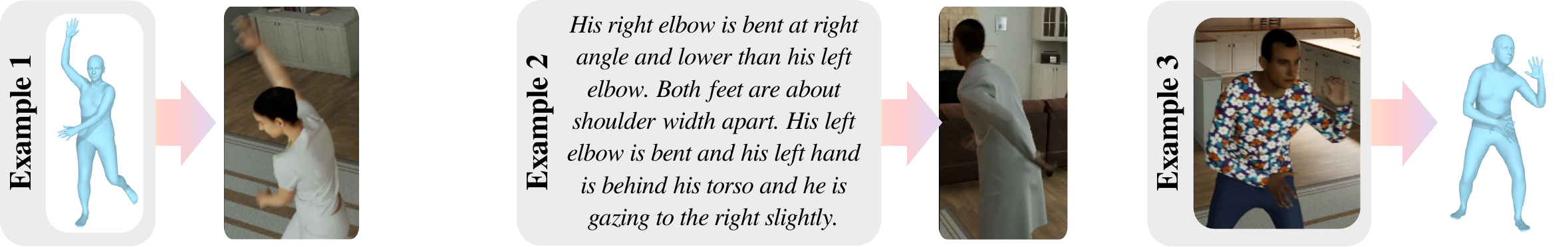} \\[-0.3cm]
    \caption{\textbf{Original paired modalities} for the queries presented in Figure 3 of the main paper (qualitative any-to-any retrieval results).}
    \label{fig:mm_retrieval_gt}
\end{figure}

The original annotations for the queries presented in Figure 3 of the main paper were not displayed alongside the results; we provide them in Figure~\ref{fig:mm_retrieval_gt}.

\section{Implementation details}
\label{app:implem_details}

\myparagraph{Architecture details.} We detail below the architecture of each of our model components:
\begin{itemize}[noitemsep,topsep=0pt,label=$\bullet$]
    \item \textit{The pose encoder} is extracted from a Variational AutoEncoder (VAE)~\cite{vae}. Its architecture follows VPoser~\cite{smplx}, and was adapted to process the first 22 SMPL-X~\cite{smplx} body joint rotations (axis-angle representation). It shares the training objectives of the pose generative model in PoseFix~\cite{delmas2023posefix}, and has been trained on the 3D poses of BEDLAM-Script. For feature representation, we use the 512-dimensional vector that is further projected in the VAE to produce the distribution parameters. This frozen pretrained representation is fed to a trainable linear layer followed by a ReLU activation.
    \item \textit{The text encoder} is the same as in~\cite{posescript}: text tokens are embedded thanks to a frozen DistilBERT~\cite{sanh2019distilbert}, then fed to a transformer~\cite{transformer} (latent dimension 512, 4 heads, 4 layers, feed-forward networks of size 1024, GELU~\cite{hendrycks2016gaussian-gelu} activations, dropout rate of 0.1). The final single-vector embedding of a text is obtained by average-pooling all its token encodings. This frozen pretrained representation is further given to a trainable linear layer followed by a ReLU activation.
    \item \textit{The image encoder} is the Vision Transformer~\cite{dosovitskiy2020image} backbone from the SMPLer-X~\cite{smplx} base model, which was connected with a neural head and trained end-to-end for human mesh recovery. This image encoder is thus assumed to yield already powerful human-aware visual features and kept frozen. While SMPLer-X reasons on all image patches at once for SMPL regression, we average pool the visual tokens during the pretraining of the PoseEmbroider, based solely on synthetic data, and do not tune them in later stages (\eg for SMPL regression, in Section 6 of the main paper). Specifically, the frozen, pretrained patch representations are aggregated into a single-vector representation after going through a trainable linear layer (projecting into a 512-dimensional space), and a ReLU activation.
    \item \textit{The Embroider model} is a transformer~\cite{transformer} (latent dimension 256, 4 heads, 4 layers, feed-forward networks of size 512, GELU~\cite{hendrycks2016gaussian-gelu} activations, dropout rate of 0.1) followed by a LayerNorm~\cite{ba2016layernorm}. It is sandwiched by two linear layers, projecting the input from a 512-dimensional space to the 256-dimensional working space of the transformer and vice-versa. Learned tokens (\ie $x$, $e_v$, $e_p$ and $e_t$) are learnable parameters of size 512. Modality-specific reprojection MLPs, processing the PoseEmbroider output $\bar{x}$, consist in small multi-layer perceptrons~\cite{haykin1994neural_mlp} with 2 fully-connected layers of size 512 and a ReLU activation in-between. Their outputs are further L2-normalized.
    \item \textit{The Aligner baseline model} appends modality-specific MLPs to each modality encoder. They consist of three linear layers with two in-between ReLU activations and dropout. Their hidden dimension is the same as the input, and they project into a 512-dimensional space.
    \item \textit{The text decoder} of the text generation model has the same architecture as in PoseFix~\cite{delmas2023posefix}, except that it takes pose encodings of size 512 instead of 32. The pose encodings are fused with TIRG~\cite{vo2019composing}, projected thanks to a linear layer of dimension 512, then fed via cross-attentions to a transformer decoder (latent dimension 512, 8 heads, 4 layers, feed-forward networks of size 1024, GELU~\cite{hendrycks2016gaussian-gelu} activations, dropout rate of 0.1), which takes 512-dimensional token encodings as input. The output tokens are given to a linear layer of the size of the vocabulary to predict the likelihood of each subsequent word.
    \item \textit{The SMPL regressor} relies on an iterative residual network as in \cite{Kanazawa_2018_CVPR}. We first concatenate the expected 512-sized input and the latest pose estimation (in the continuous 6D representation~\cite{zhou2019continuity}), then feed the outcome to a 3-layer MLP (hidden dimension: 1024), evenly stuffed with dropout (default rate of 0.5) and Leaky ReLU activations. The results is further interpreted as the next pose estimate.
    The SMPL regressor also comprises a small 2-layer MLP (hidden dimension: 512; ReLU activation), taking the 512-dimensional PoseEmbroider feature as input to predict 10 shape coefficients.
\end{itemize}

The PoseEmbroider and Aligner models have a similar size of 164.8M parameters (even though the reprojection heads of the PoseEmbroider are expandable), including 162.5M just for the encoders.

\myparagraph{Optimization and training details.}
The PoseEmbroider model is trained for 350 epochs, with all the uni-modal encoders frozen. We use mini-batches of size 128, a learning rate of $2.10^{-4}$, the Adam~\cite{kingma2014adam} optimizer and a learning rate scheduler considering steps of size 400 and a gamma value of $0.5$.

The text generation model is optimized with Adam, with a learning rate and weight decay of $10^{-4}$, for 900 epochs, and with batch sizes of 64. The finetuning on the PoseFix-OOS dataset and BEDLAM-Fix is run on 300 epochs. All trainings were done using precomputed cached features for the input pose representations.

The SMPL regression head undergoes a 100-epoch-long training with the same optimization hyper-parameters as for the text generation model.

\myparagraph{R-precision metrics for text generation.} This metric was originally proposed by \cite{chuan2022tm2t} for motion-to-text generation, and directly imported by \cite{delmas2023posefix} for instruction text generation from pose pairs. It requires to first train an auxiliary retrieval model that links annotated texts and pose pairs. Then, for each generated text, this model is used to rank a set of pose pairs. The R-precision corresponds to the maximal rank of the pose pair that was actually used to generate the text. \cite{delmas2023posefix} followed \cite{chuan2022tm2t} and used a pool size of 32, however we report results on a harder pool size of 200.

\myparagraph{Text instruction generation model: comparison with PoseFix's~\cite{delmas2023posefix}.} We explain here the differences with the setting in \cite{delmas2023posefix}, which prevents the direct comparison of the models presented in this paper with those from \cite{delmas2023posefix}. First, \cite{delmas2023posefix} trains the pose encoder from scratch alongside the text decoder, which results in a pose encoder finetuned for the studied task. In this work, as we aim to compare off-the-shelf representations and offer inference from image input, we resort to pretrained frozen (and thus potentially sub-optimal) pose encoders, however allowing a mapping to the visual space. Next, while we both use data derived from AMASS~\cite{amass} (recall that BEDLAM~\cite{Black_CVPR_2023_bedlam} uses motions from AMASS), it is not exactly the same. In particular, the construction of BEDLAM-Script and BEDLAM-Fix had to account for selection criteria on the images as well (joint visibility, inter-person occlusions, crop resolution \etc). In addition, in this work we only consider 50k poses (and 54k pairs) for pretraining, against 100k poses (and 95k pairs) in \cite{delmas2023posefix}.

Aside from the aforementioned elements, the Aligner in Table 2 of the main paper is the closest to PoseFix's original text generation model.

\section{Responsibility to human subjects}
\label{app:responsibility}

Our models were trained exclusively on synthetic data from BEDLAM~\cite{Black_CVPR_2023_bedlam} and data from PoseFix~\cite{delmas2023posefix} which includes human-written texts, but those do not carry any personal information.

The real-world images used to showcase the capabilities of our text generation model are solely used for qualitative studies. The Yoga images from Figure 6 of the main paper were obtained in studio with the written agreement of the subject. The images from Figure~\ref{fig:txt_gen_qual_nicole} were extracted from a public YouTube video, and we hide the faces to preserve anonymity.

\end{document}